\documentclass{article}

\usepackage[preprint]{neurips_2022}
\usepackage[utf8]{inputenc} 
\usepackage[T1]{fontenc}    
\usepackage{hyperref}       
\usepackage{url}            
\usepackage{booktabs}       
\usepackage{amsfonts}       
\usepackage{nicefrac}       
\usepackage{microtype}      
\usepackage{xcolor}         
\usepackage{algorithm}
\usepackage{algorithmic}
\usepackage{graphicx}

\title{Graph Value Iteration}

\author{Dieqiao Feng \\
  Department of Computer Science\\
  Cornell University\\
  \texttt{dqfeng@cs.cornell.edu} \\
  \And
  Carla P. Gomes \\
  Department of Computer Science\\
  Cornell University\\
  \texttt{gomes@cs.cornell.edu} \\
  \And
  Bart Selman \\
  Department of Computer Science\\
  Cornell University\\
  \texttt{selman@cs.cornell.edu} \\
}

\begin{document}

\maketitle

\begin{abstract}
    In recent years, deep Reinforcement Learning (RL) has been successful in various combinatorial search domains, such as two-player games and scientific discovery. However, directly applying deep RL in planning domains is still challenging. One major difficulty is that without a human-crafted heuristic function, reward signals remain zero unless the learning framework discovers any solution plan. Search space becomes \emph{exponentially larger} as the minimum length of plans grows, which is a serious limitation for planning instances with a minimum plan length of hundreds to thousands of steps. Previous learning frameworks that augment graph search with deep neural networks and extra generated subgoals have achieved success in various challenging planning domains. However, generating useful subgoals requires extensive domain knowledge. We propose a domain-independent method that augments graph search with graph value iteration to solve hard planning instances that are out of reach for domain-specialized solvers. In particular, instead of receiving learning signals only from discovered plans, our approach also learns from failed search attempts where no goal state has been reached. The graph value iteration component can exploit the graph structure of local search space and provide more informative learning signals. We also show how we use a curriculum strategy to smooth the learning process and perform a full analysis of how graph value iteration scales and enables learning.
\end{abstract}

\section{Introduction}
    
    Planning is a core area in Artificial Intelligence (AI), which emerged in the early days of AI as part of robotics research \cite{blum1997fast,hoffmann2001ff}. A planning problem consists of a start state, one or more goal states, and a set of actions that specifies how one can move from one state to the next. The generality of the planning formalism captures a surprisingly wide range of tasks, including robotic path planning, program synthesis, general theorem proving (a proof can be seen as a plan), combinatorial puzzles, etc.
    
\begin{figure*}[t]
        \centering
        \includegraphics[width=0.8\textwidth]{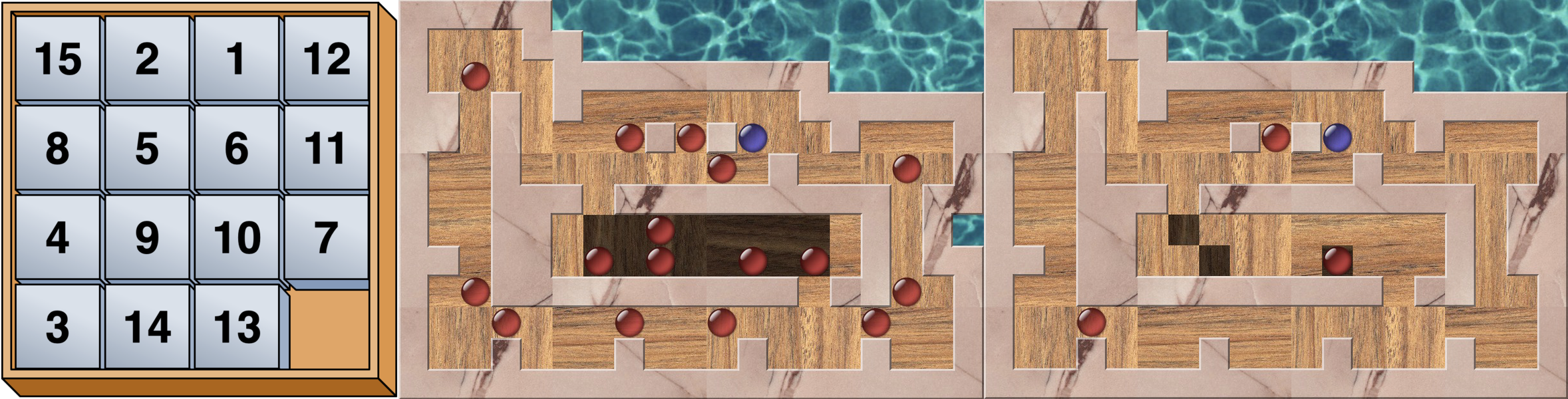}
        \caption{\textbf{Left panel:} An example of a 15-puzzle instance. \textbf{Middle panel:} An example of a Sokoban instance \protect\cite{SokobanVis2021}. The blue circle represents the player (or “pusher”), the red circles represent boxes, and the dark squares are goal squares. Light-colored squares form walls. The player has to push all the boxes onto goal squares. \textbf{Right panel:} An example of applying the sub-instance creation strategy for Sokoban used in \protect\cite{DBLP:conf/ijcai/FengGS20}. The approach randomly selects an equal number of boxes and goal squares from the original ones and leaves the locations of walls and empty squares unchanged to form a new sub-instance. The created sub-instance is easier to solve compared with the original one. In section~\ref{sec:experiment}, we show various experimental results for deep RL for Sokoban with and without sub-instances.}
        \label{fig:example}
    \end{figure*}

    In this paper, we focus on solving hard Sokoban instances that are challenging for general AI planners and even for domain-specialized solvers. In addition, we also conduct experiments on N-puzzle problems to further study the quality of discovered plans. See Figure~\ref{fig:example} for example instances for both domains. These two domains demonstrate core aspects of planning problems, e.g.,  irreversible actions vs. reversible actions, finding an arbitrary plan vs. finding a shorter plan. Both domains are representative of hard planning problems, despite their simple conceptual rules. Finding the shortest solution for N-puzzle is NP-hard and Sokoban was proved to be PSPACE-complete \cite{culberson1997sokoban,hearn2005pspace}. A regular-sized Sokoban board can have the shortest solution with hundreds or even thousands of pushes. The fact that pushes are irreversible makes solving harder. Modern N-puzzle and Sokoban solvers are based on combinatorial search augmented with carefully designed heuristic functions, pruning techniques, and various dead-end detection rules to avoid sub-search space with no solution. While we focus on Sokoban and N-puzzle problems, we stress that our approach only uses the minimum domain knowledge required to describe any planning problem, namely the state representation, the state-action transition function, and routines deciding the start state and goal states.
    
    One major difficulty for RL in planning domains is the extreme sparsity of rewards \cite{weng2020curriculum}. Unlike other domains, e.g., video games, where a well-shaped reward function can often provide immediate learning signals for each state-action pair, planning domains require a complete valid plan to generate useful learning signals. A random exploration strategy cannot ``accidentally'' encounter a valid plan consisting of hundreds of steps. \cite{DBLP:conf/nips/FengGS20,DBLP:conf/ijcai/FengGS20} proposed a curriculum approach that creates a pool of sub-instances from the input instance, selects adequate instances to try, and gradually increases the power of the framework by solving harder and harder sub-instances. Sub-instances are easier to solve and make the learning smoother. As a result, the curriculum framework can solve challenging Sokoban instances that are out of reach for domain-specialized solvers.  See the right panel of Figure~\ref{fig:example} for an example. Nevertheless, while the sub-instance creation strategy was shown to be effective for the Sokoban domain, it may not be clear how to generate such sub-instances for an arbitrary domain. So it is desirable to develop an approach that generalizes to arbitrary domains.
    
    We thus propose a graph value iteration component to enhance the learning. Traditional value iteration uses Bellman equation to update state value estimates on {\bf sampled trajectories}. For planning domains, graph search, e.g., Best-First Search (BFS) or Monte Carlo Tree Search (MCTS), provides search space with more informative combinatorial structure than trajectories, and enables value iteration to collect and exploit the information from local search graph of failed attempts. The update rule of graph value iteration is rather simple --- we update the state value estimate of each node using a variant of Bellman equation that deals with multiple outgoing edges in the reversed topological order. Compared with previous works, the graph value iteration combined with graph search and curriculum learning framework can solve more hard planning instances with less input instances, as shown in the Experiments section.
 
    \textbf{Our contributions:} We introduce a novel Graph Value Iteration (GVI) component to automatically explore and learn from planning instances with little domain knowledge. GVI caches information about the search space from each (potentially failed) search attempt and generates training data to learn the structure of local search space. Thereby the plan search network can still learn from failed search attempts, i.e., before finding a solution plan to the goal. So, even for a set of input instances where no easily solvable instance exists, we will show that GVI can still learn useful information to improve the plan search network. GVI naturally generates learning signals containing intrinsic rewards that encourage graph search to explore previously unseen states and eventually find one solution given enough curriculum iterations. We show that our approach allows us to solve substantially harder instances than previous works with less input instances.

\section{Formal Framework}
    \begin{figure*}[t]
        \centering
        \includegraphics[width=0.8\textwidth]{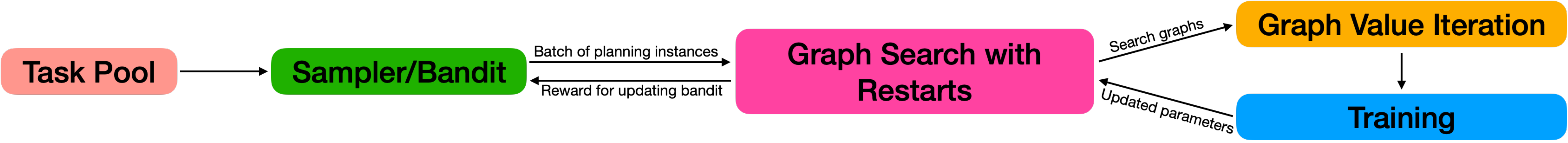}
        \caption{\textbf{The main workflow.} The input of the framework is a task pool consisting of multiple planning instances to solve, and the output after learning is the set of solved instances with the shortest found solutions attached. In each iteration, the sampler picks a batch of planning instances from the task pool, and the graph search component, which is BFS or MCTS, attempts to solve these instances. The success/failure status of each instance is then sent back to the sampler to adjust its weights. A graph value iteration is performed on each search graph generated by graph search to provide further training data for the training phase. The updated parameters are then sent back to the search component and a new iteration starts.}
        \label{fig:workflow}
    \end{figure*}
    
    An overview of the workflow of one full iteration of the curriculum framework is depicted in Figure~\ref{fig:workflow}. The framework aims to solve hard planning instances (namely Sokoban or 15-puzzle instances) with no labeled solutions given. Subsection~\ref{sec:data_gen} describes the details about how to generate learning signals from local search graph, which is the \emph{core novel component} of our approach.

    \subsection{Curriculum Learning Framework and Training Process}
    \label{sec:curriculum}
    We adapted the curriculum strategy of \cite{DBLP:conf/nips/FengGS20} for our approach. The input to the learning framework is a set of planning instances to solve, and after learning the output contains solved instances with the shortest found solutions attached. In each iteration, we randomly sample a batch of input instances and run the search on each instance with a fixed search budget. Training data is then collected from the search graph to improve the parameters of the network. Initially, the model can only solve the easiest instances of the input set. As the learning goes, the prediction of the network gradually becomes more accurate and the model can solver harder instances. And training data collected from solved hard instances can further improve the network and this procedure iterates until the model cannot solve any more new instances and terminates. This offers a natural curriculum view of the input planning instances.
    
    For training, each batch samples state-action pairs from discovered plans and those generated by GVI. Specifically, labels for state $s_i$ on a discovered plan $(s_0,a_0,...,a_{n-1},s_n)$ is set to $n - i$ and labels for GVI-generated learning signals are described in Algorithm~\ref{alg:data_gen}. The percentage of GVI-generated samples in each batch is control by a constant $p$. In Section~\ref{sec:ab} we do an ablation test to see how $p$ affects the overall performance.

    \subsection{Graph Search with Restarts}
    \label{sec:astar}
    
    Our search component is an augmentation of vanilla graph search. Specifically, instead of using a manually crafted heuristic function, a deep neural network $v = h_\theta(s)$ with parameters $\theta$ takes a state $s$ as input and predicts a scalar value $v$ indicating the estimated remaining steps to the closest goal state. We compare two search methods, namely BFS and MCTS, in this paper.

    Given any start state $s$, BFS tries to find solution paths from $s$ to any goal state. The search maintains an open set which contains boundary nodes that are not expanded. At each search step of the main loop, BFS determines the next node to expand from the open set by picking the node with minimum evaluation function $f(n)$. We adopt the evaluation function of weighted A\textsc{*} and set
    \[f(n) = g(n) + 2 \cdot h_\theta(\textrm{state of } n),\]
    where $g(n)$ is the cost from the start node to $n$ \cite{pohl1970first}. Notice that unlike A\textsc{*} which guarantees optimality and requires the heuristic function to be admissible and consistent, our learning objective is to shape the heuristic predictor as close as possible to the ground truth. Thus discovered solution plans by BFS are not guaranteed to be optimal.
    
    MCTS tries to find the next best action given any state $s$. Since the output of the network $h_\theta(s)$ is the remaining distance estimate, we use $r(s) = -\log(h_\theta(s))$ as the reward of leaf nodes. At each selection stage, MCTS selects the next node with maximum
    \[
        U(s, a) = Q(s, a) + \sqrt{\frac{2 \cdot \ln \sum_b N(s, b)}{N(s, a)}},
    \]
    where $N(s, a)$ is the visit count of the action $a$ on the state $s$ and $Q(s, a)$ is the average rewards from previous simulations.

    We use graph search with \textit{restarts}~\cite{Biere2018,restart}. For each planning instance, instead of performing a long search, we divide the search budget into small pieces and each BFS run can only expand at most $S$ nodes. Multiples short runs can avoid exponential mean runtime when the runtime distribution has heavy-tailed behavior~\cite{restart}.

    \subsection{Dataset and Graph Value Iteration}
    \label{sec:data_gen}

    \noindent The study of the performance of search methods is greatly hampered by the difficulty in collecting real data and using procedurally generated instances may result in too regular search space. To make our result more diverse and general, we collected the Large Test Suite Dataset from the Sokobano website\footnote{http://sokobano.de/en/levels.php}, resulting in 3272 instances in total. All these instances were designed by different human authors in the past few decades, have great variation in the underlying structure, serve as the benchmark for specialized solvers, and also exhibit practical interest for humans to solve.

    \begin{figure*}[t]
        \centering
        \includegraphics[width=0.6\textwidth]{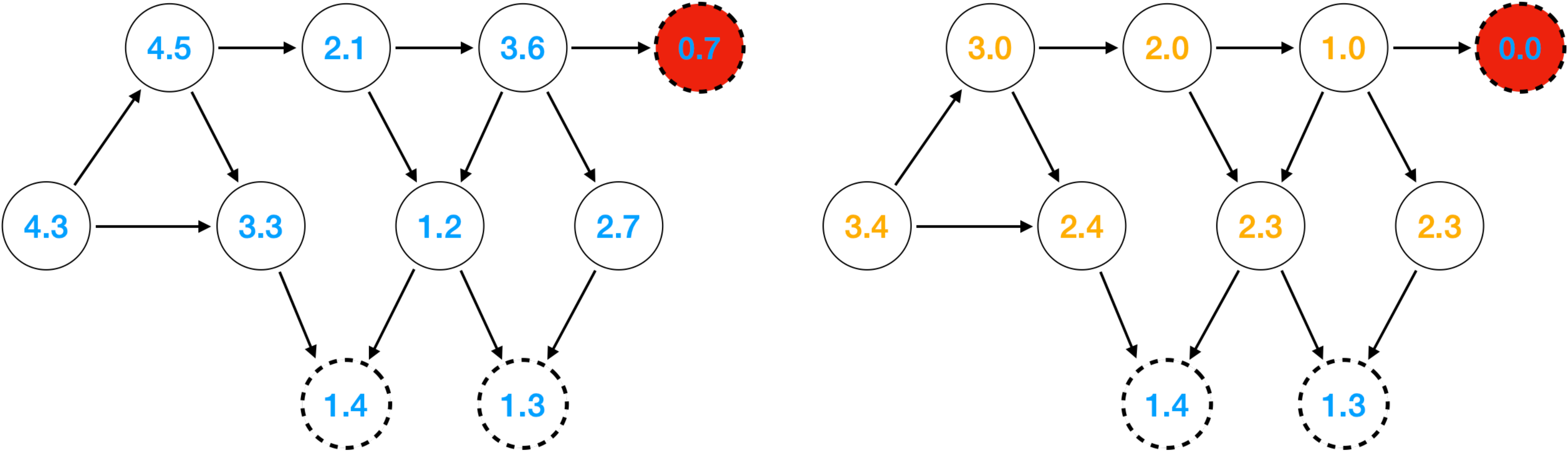}
        \caption{\textbf{The use of graph value iteration to calculate new training labels.} \textbf{Left panel:} The graph generated by graph search. \textbf{Right panel:} The graph with adjusted training labels. Solid circles represent closed nodes, dashed circles represent open nodes, and red circles are found goal states. Blue labels are remaining distance predicted by the network and yellow labels are calculated using graph value iteration. Each time the search hits the maximum node expansions, we fix the predicted labels of open nodes (found goal states are considered as open nodes too with labels modified to zeros) and calculate the labels of the remaining graph according to a variant of Bellman equation in the reversed topological order (see Algorithm~\ref{alg:data_gen} for more details). The calculated yellow labels reflect information about the local search structure near the start state and are then used as training signals.}
        \label{fig:dijkstra}
    \end{figure*}
    
    \begin{algorithm}[t]
        \caption{The Graph Value Iteration Component}
        \label{alg:data_gen}
        \textbf{Input}: Search graph $\mathcal{G}$ generated by graph search, open set $\mathcal{O}$
        \begin{algorithmic}[1]
            \STATE Let $\mathcal{U} = \textrm{the whole node set of }\mathcal{G}$.
            \FOR{$u\in \mathcal{U}$}
                \IF{$u$ is a goal state}
                    \STATE $d(u) = 0$ and remove $u$ from $\mathcal{O}$.
                \ELSIF{$u\in\mathcal{O}$}
                    \STATE $d(u) = h_\theta(s(u))$ (raw network prediction of the state $s(u)$.
                \ELSE
                    \STATE $d(u) = \infty$\ (set infinite to all closed nodes).
                \ENDIF
            \ENDFOR
            \WHILE{$\mathcal{U} \neq \emptyset$}
                \STATE Let $v = \textrm{arg\,min}_u d(u)$.
                \FOR{each node $w$ in incoming neighbors of $v$}
                    \STATE Let $d(w) = \min(d(w), d(v) + 1)$.
                \ENDFOR
                \STATE Let $\mathcal{U} = \mathcal{U} \setminus \{v\}$.
            \ENDWHILE
            \STATE \textbf{return} each node $u$ that are not in $\mathcal{O}$ with label $d(u)$
        \end{algorithmic}
    \end{algorithm}

    Unlike previous works that only collect training data from found plans, we generate training data from both successful and failed search attempts. When a failed search has been done, we fix the deep net's predictions of nodes from the open set (boundary/leaf nodes) and then perform a graph value iteration to calculate the adjusted heuristics of the remaining search graph. For each goal state, we force its heuristic to be zero instead of using the predicted value. See Figure~\ref{fig:dijkstra} and Algorithm~\ref{alg:data_gen} for more details. We will show that data generated from failed search attempts can behave as curiosity-driven intrinsic rewards that encourage the search to explore previously nonvisited states.

    \subsection{Network Architecture}
    \label{sec:network}
    The planning domains considered in this paper are all board games and we used ResNet \cite{he2016deep} as the feature extractor. Commonly, different planning instances have varying board sizes. We adopt a masking strategy used in \cite{wu2019accelerating} that allows mixing multiple board sizes into the same batch during both training and evaluating. Specifically, we size each input tensor to the size of the largest height and width in the batch and zero-pad each entry independently in the batch to fill out the excess space. The network also takes 0-1 mask as input indicating which parts of the tensor are real entries instead of excess space. For each network operation that will make excess space non-zero, we multiple intermediate tensors with the mask to make excess space zero again. Special attention is needed for operations involving global statistics to ensure only real entries are taken into account.

\section{Experiments}
    \label{sec:experiment}

    The setup of all experiment runs takes a set of planning instances as input and terminates until no new instance of the input set can be solved for 10 iterations. For BFS, node expansion budget $S$ was set to 20000 for Sokoban and 500 for N-puzzles. For MCTS, we use 100 simulations for both domains. Each experiment run takes at most 3 days with 5 Nvidia Tesla V100 16GB GPUs and 2x18 core Xeon Skylake 6154 CPUs with base clock 3GHz.
    
    \subsection{Experiment Results on Sokoban}
    \label{sec:data_eff}
    
     \begin{figure}[t]
        \centering
        \includegraphics[width=\linewidth]{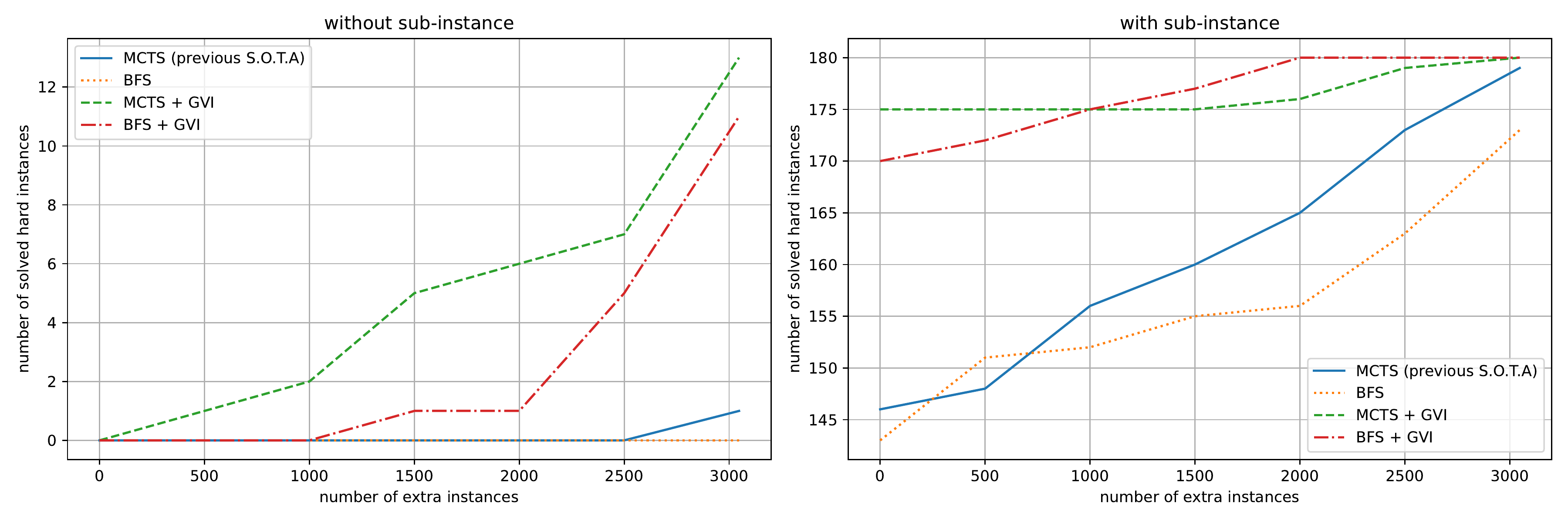}
        \caption{\textbf{Scaling comparison between MCTS and BFS with or without GVI on a set of hard Sokoban instances.} Figure~\ref{fig:example} illustrates the sub-instance creation strategy. \protect\cite{DBLP:conf/nips/FengGS20} focused on a set of hard instances that are not solvable with any other method. They collected 180 of such instances. Without sub-instance creation, both MCTS and BFS can only solve at most one hard instance. While enhanced by GVI, the number of solved instances raises to $>10$. With the sub-instance creation strategy, \protect\cite{DBLP:conf/nips/FengGS20} could solve 146 using MCTS. The remaining 34 unsolved instances could only be solved by adding almost 3000 other easier Sokoban instances into the input set. While enhanced by GVI, even without any extra instance, both MCTS and BFS can solve $\geq 170$ hard instances, which shows GVI is highly data-efficient.}
        \label{fig:compare}
    \end{figure}
    
    In \cite{DBLP:conf/nips/FengGS20}, the framework based on MCTS can solve 180 extremely hard Sokoban instances from the Large Test Suite \cite{SokobanRep2020} with the sub-instance creation strategy and extra input instances. This is a remarkable result since modern fine-tuned specialized Sokoban solvers cannot solve any of them. Introducing extra instances or creating sub-instances can greatly increase the number of solvable hard instances because more instances can prevent overfitting, smooth the whole learning procedure, and fill the ``difficulty gap'' between hard instances. In the first experiment, we disable the usage of sub-instance creation since it utilizes extensive domain knowledge. We gradually added more remaining instances from the Large Test Suite to the input set to test how many hard instances the algorithm can solve as the number of input instances increases. In the second experiment, we enabled the usage of sub-instances to match the result listed in \cite{DBLP:conf/nips/FengGS20}. See Figure~\ref{fig:compare} for more details.

    MCTS (\cite{DBLP:conf/nips/FengGS20}) cannot solve any hard instance without the sub-instance creation strategy --- thousands of extra instances are not enough to fill the ``difficulty gap''. In contrast, MCTS + GVI can solve 13 hard instances given the whole Large Test Suite as input. With the help of sub-instance creation, the overall impact of GVI is marginal given enough input instances. However, GVI still solves more instances when only a few input instances are available. Moreover, since in many practical applications one does not have collections of thousands of (non-random) problem instances, the data efficiency of GVI offers a significant advantage.
    
     \begin{table}[t]
        \centering
        \begin{tabular}{|l|l|l|l|l|l|l|}
            \hline
            Number of input instances & 500 & 1000 & 1500 & 2000 & 2500 & 3000 \\
            \hline
            MCTS (previous S.O.T.A) & 417 & 810 & 1190 & 1513 & 1937 & 2371 \\
            \hline
            MCTS + GVI & 471 & 923 & 1310 & 1792 & 2104 & 2763 \\
            \hline
            BFS & 217 & 629 & 1037 & 1501 & 1901 & 2316 \\
            \hline
            BFS + GVI & 297 & 817 & 1207 & 1561 & 1998 & 2413 \\
            \hline
        \end{tabular}
        \caption{Number of solved instances based on MCTS and BFS with or without GVI.}
        \label{tab:gvi}
    \end{table}
    Besides the hard instances, we test the performance of GVI on all 3272 instances with varying difficulties. As shown in Table~\ref{tab:gvi}, GVI can consistently improve the number of solved instances for both MCTS and BFS, regardless the number of input instances.
    
    \subsection{The Ablation Test of GVI}
    \label{sec:ab}
     \begin{table}[t]
        \centering
        \begin{tabular}{|l|l|l|l|l|l|l|l|}
            \hline
            $p$ & $0\%$ & $20\%$ & $40\%$ & $60\%$ & $80\%$ & $90\%$ & $100\%$ \\
            \hline
            MCTS (previous S.O.T.A) & 146 & 156 & 171 & {\bf 175} & 163 & 97 & 0 \\
            \hline
            BFS & 143 & 147 & {\bf 170} & 164 & 117 & 83 & 0 \\
            \hline
        \end{tabular}
        \caption{Number of solved hard instances according to the percentage $p$ of training samples from GVI. The table shows the best performance is reached when mixing learning signals from discovered plans and GVI.}
        \label{tab:ablation}
    \end{table}
    In this section, we do an ablation test of GVI to test the overall performance based on the weight factor of GVI. Specifically, each batch of the training procedure samples state-value pairs from discovered plans and those generated by GVI from failed search attempts. We can control the percentage $p$ of GVI samples in one batch to test how it affects the learning. Notice that $p=0$ means no GVI while $p=1$ means that learning signals purely come from failed attempts. We enables sub-instance creation and disables extra instance to show the impact of GVI. See Table~\ref{tab:ablation} for more details.
    
    Without GVI, both BFS and MCTS can solve around 145 hard instances. The number of solved instances increases when we start to mix in the learning signals from GVI, and drops to 0 when using pure GVI.

    \subsection{How Pure GVI Shapes the Local Search Space}
    \label{sec:no_sol}
    
       \begin{figure*}[t]
        \centering
        \includegraphics[width=0.9\textwidth]{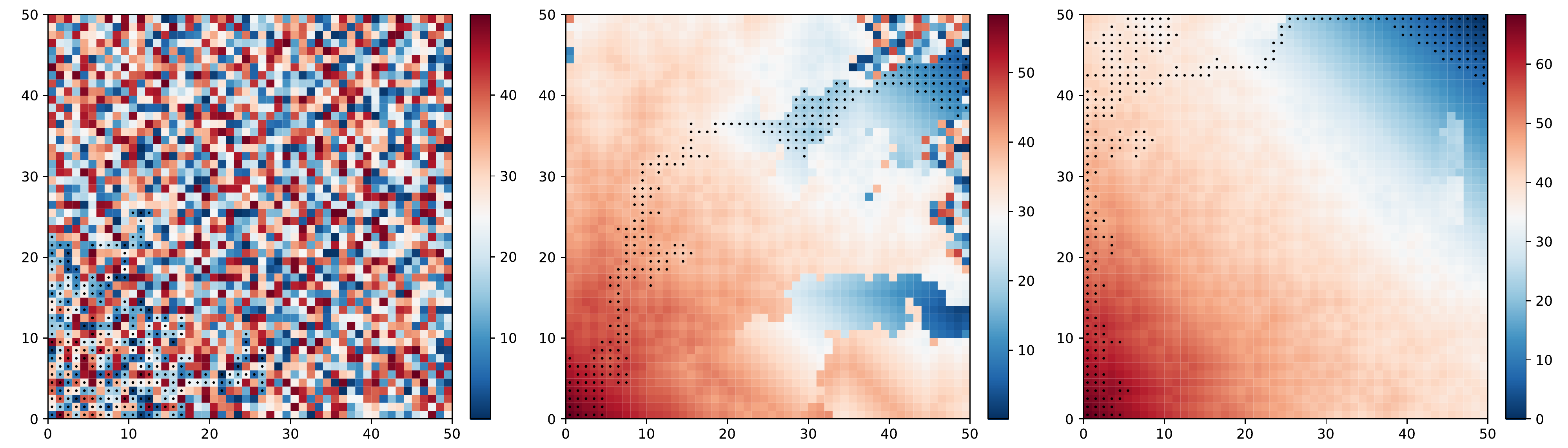}
        \caption{\textbf{Exploration behavior of best-first search + GVI with all failed attempts.} The three panels represent data collected from learning iteration 1, 50, and 100, respectively. In a 50x50 board, the robot starts at the bottom-left corner and tries to reach the goal cell located at the top-right corner. For each iteration, the maximum node expansion of the search is 300. The estimated remaining distance to the goal is represented by the color of each cell and is initialized by a uniform distribution $\mathcal{U}(0, 50)$. Cells with a centered black dot represent expanded nodes by the search at the corresponding iteration. Initially, best-first search behaves similarly to breadth-first search, and expanded states are all near the start state. As the learning goes, with learning signals from GVI, the search picks a path quickly leading to the boundary of previously visited cells and starts exploring from there. The middle panel shows the search is exploring two areas of sub-space simultaneously (blue areas).}
        \label{fig:explore}
    \end{figure*}
    
    Herein, we show that pure GVI can still learn usable local search structure even without any solution found. To demonstrate how learning shapes local search structure, we construct a toy setting with a 50x50 grid board and the goal for the robot is to reach the upper-right corner from the lower-bottom corner. For each step, the robot can move to one of its four adjacent cells and the maximum search steps for best-first search is set to 300, which is one magnitude smaller than the total number of cells. To simplify and demonstrate how learning goes, we use a table instead of a deep network to store remaining distance predictions. All heuristic values are initialized by a uniform distribution $\mathcal{U}(0, 50)$. Figure~\ref{fig:explore} shows the learning procedure. In the first iteration, the search degrades to breadth-first search and explores local cells near the start cell. After learning the local structure near the start cell, the search picks a path that leads to the boundary of previously visited cells and starts exploring from there. The middle panel of Figure~\ref{fig:explore} shows best-first search is exploring two areas of sub-search-spaces (colored with a blue cluster of cells) simultaneously. After the first time the search reaches the goal cell, the path leading to the goal cell is quickly solidified and repeatable in future iterations as shown in the right panel.
    
    \subsection{High Quality of Discovered Solutions}
    \label{sec:n_puzzle}
    
    \begin{figure*}[t]
        \centering
        \includegraphics[width=0.9\textwidth]{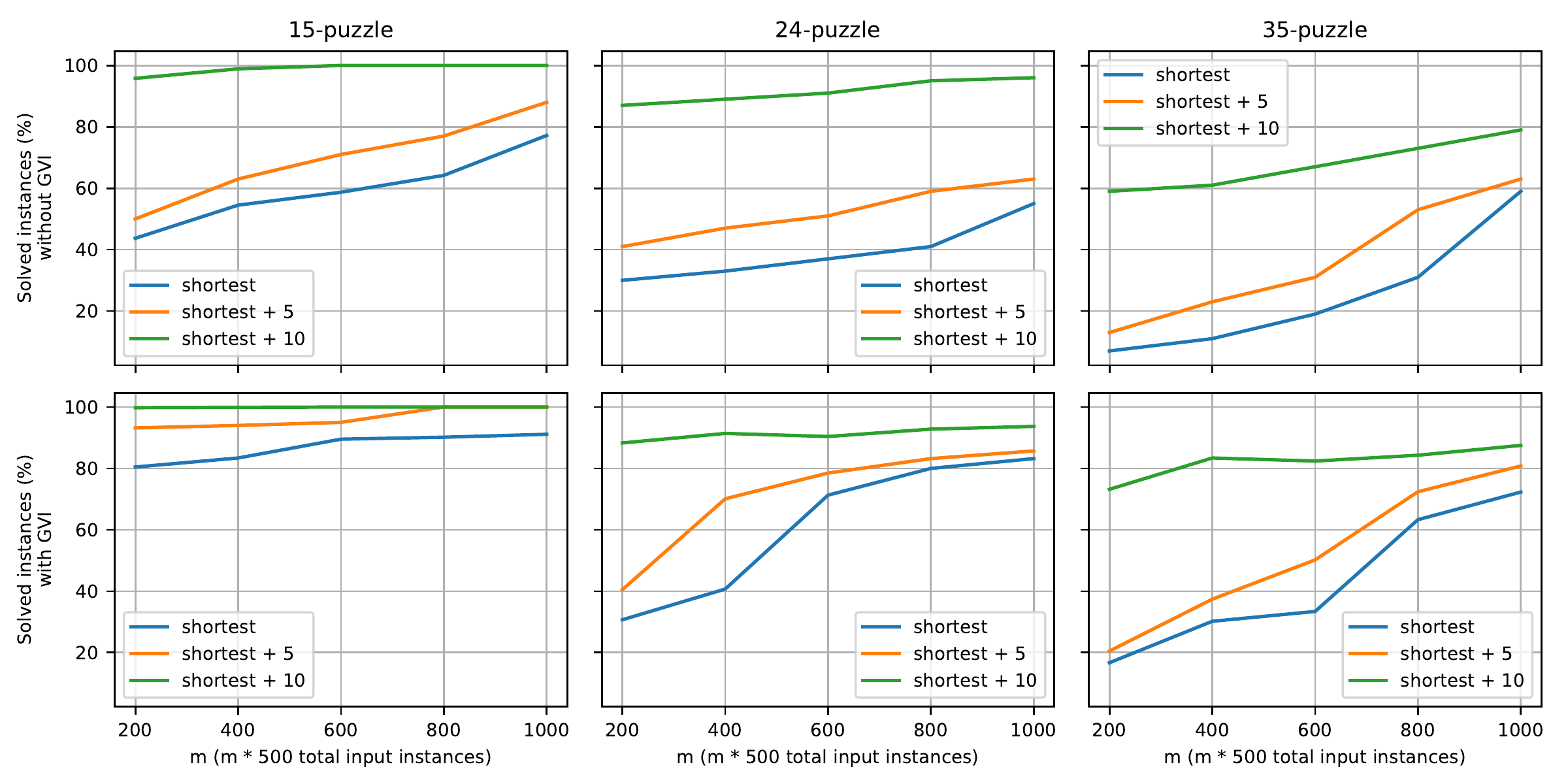}
        \caption{\textbf{Percentage of solved instances as the size of input instances increases}. Horizontal axes represent different $m$, ranging from 200 to 1000. For each $m$ we generated 500 boards with different difficulty, which were obtained by randomly scrambling $k$ times a goal board ($k=1,...,500$). The lower the $k$, the ``easier'' the board. So, our setting involves training datasets ranging from $200 \times 500=10,000$ to $1,000 \times 500=500,000$ input instances. Vertical axes show the percentage of all input instances solved by our algorithm. Different curves show the percentage of solutions found with different length relaxation compared with the shortest solutions (ground truth) searched by the optimal N-puzzle solver. For example, the curve ``shortest + 5'' means the percentage of solved instances whose found plan has length less than or equal to the ground truth length plus five. The figure shows that even with a network-generated heuristic function that is neither admissible nor consistent, solutions found by GVI still have higher quality than non-GVI 
        BFS. Notice that the overall solvability ratio increases as the number of input instances increases.}
        \label{fig:n_puzzle}
    \end{figure*}
    
    A\textsc{*} is guaranteed to find the shortest solution if the heuristic function is admissible and consistent. Herein we show that even with a network-predicted heuristic function, solutions found by GVI are still close to the shortest ones, and adding GVI can greatly improve the quality of found plans compared with the framework without GVI. We test on 15-puzzle, 24-puzzle, and 35-puzzle since optimal solvers for these board sizes exist \cite{felner2004additive}. For an input set of N-puzzle instances, we first calculated the length of the shortest solution for each instance as the ground truth. We did not prune any node whose depth exceeds the shortest length to avoid leaking ground truth information and making the search easier. We calculated the percentage of the input instances whose found shortest plans are exactly the ground truth, or do not exceed the ground truth plus a certain constant value. To generate the set of input instances, for each $k=1,...,500$, we generate $m$ different boards by randomly scrambling the goal state $k$ times. We only test on BFS for this experiment setting. See Figure~\ref{fig:n_puzzle} for more details.

    As shown in Figure~\ref{fig:n_puzzle}, even without any hint about the length of shortest solutions, solutions found by GVI still have higher quality than non-GVI BFS. For each instance of 15-puzzle, GVI can find solutions of length no longer than the ground truth plus ten. The number of input instances greatly affects the overall percentage of solved instances. This result shows the potential of GVI since generating more planning instances is cheap (we do not need to provide any other information, e.g., ground truth solutions) and in many planning domains, a large set of unlabeled instances is available.

    \subsection{Exploration Efficiency Comparison}
    \label{sec:exp_eff}
    
    \begin{table}[t]
        \centering
        \begin{tabular}{|l|l|l|l|l|l|}
            \hline
            & Time & Iteration & Node & Unique node \\
            \hline
            MCTS & 61 mins & 197 & 7674642 & {\bf 1307017}  \\
            \hline
            BFS & {\bf 35 mins} & {\bf 153} & {\bf 1457029} & 1339486  \\
            \hline
        \end{tabular}
        \caption{Resource efficiency between BFS and MCTS. We use a hard Sokoban instance, XSokoban\_29, as the test instance. Both experiments use the sub-instance creation strategy and terminate once the solution for XSokoban\_29 is found. The number of uniquely visited nodes of both algorithms is similar, but MCTS expands significantly more repreated nodes than BFS. This is mainly caused by the simulation mechanism of MCTS that visits branches under the root  node multiple times.}
        \label{table1}
    \end{table}
    \noindent In this section, we compare the efficiency of using resources between BFS and MCTS when both are augmented with GVI. For AlphaZero-style frameworks, self-plays take the majority of the time because MCTS needs to spend multiple simulations to find the best move and each node will be visited multiple times during the construction of Monte Carlo trees. In contrast, best-first search visits each node at most once and the search only needs to query a min-heap to find the best node to expand. We pick XSokoban\_29 as our test instance. See Table~\ref{table1} for more details. We found the number of unique nodes visited is roughly the same to solve the original instance. However, MCTS visits more repeated nodes because of its simulation mechanism. The total running time of MCTS is still close to BFS mainly because MCTS uses a hash table to save previously estimated boards. Whenever a repeated board is encountered, the saved estimate is fetched instead of evaluating the network. However, for larger and more complex planning instances such hashing method will have huge memory usage. For BFS, the numbers of visited nodes and uniquely visited nodes are very close, indicating the high efficiency of node expansion.
    
\section{Related Work}
    The recent breakthrough of deep RL on two-player games, e.g, Go and Chess, which are also discrete combinatorial tasks, shed a light on using deep RL for AI planning \cite{silver2017mastering}. Multiple methods have been proposed to tackle hard planning problems with deep neural networks. \cite{trunda2020deep,yonetani2020path} used a supervised method that learns the heuristic function from a set of solved planning instances. \cite{DBLP:conf/ijcai/FengGS20,DBLP:conf/nips/FengGS20} used a curriculum framework with sub-instances and solved challenging Sokoban instances that specialized solvers cannot solve. \cite{crippa2022analysis} compared MCTS and IDA\textsc{*} on Sokoban and has shown that IDA\textsc{*} provides the best performance. \cite{shoham2021solving} augmented traditional RL with backward search and solved 88 of the 90 XSokoban levels, which is the default benchmark for Sokoban. \cite{czechowski2021subgoal} used an approach of generating $k$-th step ahead subgoals on Sokoban. However, the method requires an expert dataset which consists of suboptimal solutions. \cite{agostinelli2021search} used Deep Approximate Value Iteration (DAVI) to learn a heuristic function for Rubik's cube in an unsupervised manner. States for training are generated by randomly backward scrambling the goal state $k$ times. However, for hard planning domains that require hundreds of steps to solve, pure random backward scrambling cannot generate challenging enough states. 
    \cite{graves2017automated} proposed the idea of \emph{automatic curriculum learning} to learn an adaptive policy on a N-task curriculum. \cite{pathak2017curiosity,burda2018large} used prediction error as an intrinsic reward to help the agent explore the search space more efficiently. The method achieves high scores on multiple Atari games purely guided by curiosity-driven rewards. GoExplore further achieved significant score improvement on two hard-exploration domains of Atari: Montezuma's Revenge and Pitfall \cite{ecoffet2019go}. GoExplore remembers states that have previously been visited, return to a promising state (without exploration), and then explore from it.
    
    Herein we listed three main features GVI offers: 1) it doesn't require any expert dataset; 2) it can solve challenging Sokoban instances that are out of reach for specialized solvers and other approaches; 3) it is data-efficient. Without sub-instance creation, GVI can scale up way better than previous MCTS-based methods.

\section{Conclusions}
\label{sec:conclusions}
    Recently, there has been significant progress on using deep RL to solve highly challenging AI planning problems in the Sokoban domain with over thousand-step solution plans \cite{DBLP:conf/ijcai/FengGS20}. The curriculum framework learns from a collection of easier sub-instances. Training is effective because the simpler instances can be solved with MCTS, which provides useful reward signals to train a deep net incrementally to guide the search for solving harder instances.

    Here we show how we can also learn without sampling sub-instances but from other problem instances available in problem benchmark sets. Most importantly, we introduce a mechanism for learning search guidance even when the search does not yield any valid plans (i.e., no goal state is reached). Our graph value iteration algorithm (\ref{fig:dijkstra}) provides learning signals about the heuristic function (estimating the distance to the goal) that is derived from the structure of the local search space.

    Our experiments show that there is useful information in the updated heuristic function, even before a problem instance is solved. This opens up a promising new direction in general for deep RL on hard combinatorial planning domains where true rewards (reaching a goal state) can be thousands of steps away. In such settings, the probability for MCTS with random roll-outs to accidentally find a solution path is too small. However, GVI can learn an approximate search policy long before reaching the goal state, which is in turn used to boost the remaining search.

\clearpage
\bibliography{neurips_2022}


\end{document}